\documentclass[10pt,twocolumn,letterpaper]{article}

\usepackage{wacv}

\makeatletter
\@namedef{ver@everyshi.sty}{}
\makeatother
\usepackage{tikz}

\usepackage{times}
\usepackage{epsfig}
\usepackage{graphicx}
\usepackage{amsmath}
\usepackage{amssymb}
\usepackage{booktabs}

\def\wacvPaperID{1398} 

\wacvalgorithmstrack   

\wacvfinalcopy 

\ifwacvfinal
\usepackage[breaklinks=true,bookmarks=false]{hyperref}
\else
\usepackage[pagebackref=true,breaklinks=true,colorlinks,bookmarks=false]{hyperref}
\fi

\usepackage[noend]{algpseudocode}
\usepackage{subfigure, multirow}
\usepackage{booktabs}
\usepackage{array}
\usepackage{amsmath}
\usepackage{diagbox}
\usepackage{xspace}
\usepackage{cleveref}
\usepackage{bm}
\usepackage[ruled,linesnumbered,vlined]{algorithm2e}
\usepackage{graphicx}
\usepackage{floatrow}
\newfloatcommand{capbtabbox}{table}[][\FBwidth]
\usepackage{pgfplots}
\usepackage{wrapfig}
\usepackage{ragged2e}
\usepackage{xcolor}

\newcommand{\bx}{\bm{x}}

\newcommand{\bg}{\bm{g}}
\newcommand{\bs}{\bm{s}}

\newcommand{\bw}{\bm{w}}

\newcommand{\be}{\bm{e}}
\newcommand{\btheta}{\bm{\theta}}
\newcommand{\bvartheta}{\bm{\vartheta}}

\newcommand{\mt}{\mathcal{T}}
\newcommand{\ms}{\mathcal{S}}
\newcommand{\ml}{\mathcal{L}}
\newcommand{\ma}{\mathcal{A}}

\makeatletter
\DeclareRobustCommand\onedot{\futurelet\@let@token\@onedot}
\def\@onedot{\ifx\@let@token.\else.\null\fi\xspace}
\def\eg{\emph{e.g}\onedot} 
\def\ie{\emph{i.e}\onedot}

\def\etal{\emph{et al}\onedot}
\renewcommand{\paragraph}{%
	\@startsection{paragraph}{4}{\z@}%
	{0.1em \@plus 0.5ex \@minus 0.2ex}{-1em}%
	{\normalsize\bf}%
}
\makeatother

\usepackage{amsmath,amsfonts,bm}

\def\eqref#1{equation~\ref{#1}}

\def\1{\bm{1}}

\DeclareMathAlphabet{\mathsfit}{\encodingdefault}{\sfdefault}{m}{sl}
\SetMathAlphabet{\mathsfit}{bold}{\encodingdefault}{\sfdefault}{bx}{n}

\DeclareMathOperator*{\argmin}{arg\,min}

\pgfplotsset{compat=1.17} 

\begin{document}

\title{Dataset Condensation with Distribution Matching}

\author{{Bo Zhao}, {Hakan Bilen}\\
School of Informatics, The University of Edinburgh\\
{\tt\small \{bo.zhao, hbilen\}@ed.ac.uk}
}

\maketitle
\thispagestyle{empty}

\begin{abstract}
Computational cost of training state-of-the-art deep models in many learning problems is rapidly increasing due to more sophisticated models and larger datasets. A recent promising direction for reducing training cost is dataset condensation that aims to replace the original large training set with a significantly smaller learned synthetic set while preserving the original information. While training deep models on the small set of condensed images can be extremely fast, their synthesis remains computationally expensive due to the complex bi-level optimization and second-order derivative computation. In this work, we propose a simple yet effective method that synthesizes condensed images by matching feature distributions of the synthetic and original training images in many sampled embedding spaces. Our method significantly reduces the synthesis cost while achieving comparable or better performance. Thanks to its efficiency, we apply our method to more realistic and larger datasets with sophisticated neural architectures and obtain a significant performance boost\footnote{The implementation is available at \url{https://github.com/VICO-UoE/DatasetCondensation}.}. We also show promising practical benefits of our method in continual learning and neural architecture search. 
\end{abstract}

\section{Introduction}


Computational cost for training a single state-of-the-art model in various fields, including computer vision and natural language processing, doubles every 3.4 months in the deep learning era due to larger models and datasets.
The pace is significantly faster than the Moore's Law that the hardware performance would roughly double every other year~\cite{AIandCompute}. 
While training a single model can be expensive, designing new deep learning models or applying them to new tasks certainly require substantially more computations, as they involve to train multiple models on the same dataset for many times to verify the design choices, such as loss functions, architectures and hyper-parameters \cite{bergstra2012random, elsken2019neural}.
For instance, Ying \emph{et al.} \cite{ying2019bench} spend 100 TPU years of computation time conducting an exhaustive neural architecture search on CIFAR10 dataset \cite{krizhevsky2009learning}, while training the best-performing architectures take only dozens of TPU minutes. 
Hence, there is a strong demand for techniques that can reduce the computational cost for training multiple models on the same dataset with minimal performance drop. 
To this end, this paper focuses on lowering the training cost by reducing the training set size.

The traditional solution to reduce the training set size is coreset selection. Typically, coreset selection methods choose samples that are important for training based on heuristic criteria, for example, minimizing distance between coreset and whole-dataset centers \cite{chen2010super, rebuffi2017icarl, castro2018end, belouadah2020scail}, maximizing the diversity of selected samples \cite{aljundi2019gradient}, discovering cluster centers \cite{farahani2009facility, sener2017active}, counting the mis-classification frequency \cite{toneva2019empirical} and choosing samples with the largest negative implicit gradient \cite{borsos2020coresets}. Although coreset selection methods can be very computationally efficient, they have two major limitations. First most methods incrementally and greedily select samples, which are short-sighted. 
Second their efficiency is upper bounded by the information in the selected samples in the original dataset. 

An effective way of tackling the information bottleneck is \emph{synthesizing} informative samples rather than selecting from given samples.
A recent approach, dataset condensation (or distillation)~\cite{wang2018dataset, zhao2021DC} aims to learn a small synthetic training set so that a model trained on it can obtain comparable testing accuracy to that trained on the original training set.
Wang \emph{et al.} \cite{wang2018dataset} pose the problem in a learning-to-learn framework by formulating the network parameters as a function of synthetic data and learning them through the network parameters to minimize the training loss over the original data.
An important shortcoming of this method is the expensive optimization procedure that involves updating network weights for multiple steps for each outer iteration and unrolling its recursive computation graph.
Zhao \emph{et al.} \cite{zhao2021DC} propose to match the gradients w.r.t. the network weights giving real and synthetic training images that successfully avoids the expensive unrolling of the computational graph.
Another efficiency improvement is a closed form optimizer by posing the classification task as a ridge regression problem to simplify the inner-loop model optimization \cite{bohdal2020flexible, nguyen2021dataset}.
In spite of the recent progress, the dataset condensation still requires solving the expensive bi-level optimization which jeopardizes their goal of reducing training time due to the expensive image synthesis process.
For instance, the state-of-the-art \cite{zhao2021DSA} requires 15 hours of GPU time to learn 500 synthetic images on CIFAR10 which equals to the cost of training 6 deep networks on the same dataset.
In addition, these methods also require tuning multiple hyper-parameters, \eg the steps to update synthetic set and network parameters respectively in each iteration, that can be different for different settings such as sizes of synthetic sets. 

\begin{figure}[]
    \centering
    \includegraphics[width=0.95\linewidth]{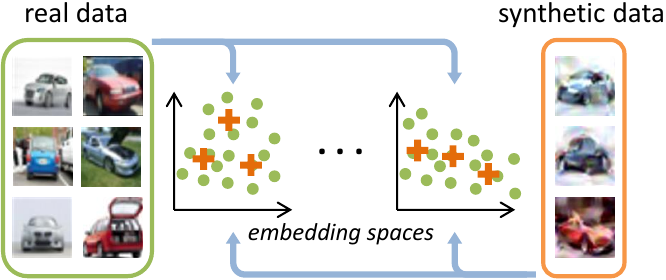}
    \caption{{Dataset Condensation with Distribution Matching. We randomly sample real and synthetic data, and then embed them with the randomly sampled deep neural networks. We learn the synthetic data by minimizing the distribution discrepancy between real and synthetic data in these sampled embedding spaces.}}
    \label{fig:method}
\end{figure}

In this paper, we propose a novel training set synthesis technique that combines the advantages of previous coreset and dataset condensation methods while avoiding their limitations.
Unlike the former and like the latter, our method is not limited to individual samples from original dataset and can synthesize training images.
Like the former and unlike the latter, our method can very efficiently produce a synthetic set and avoid expensive bi-level optimization.
In particular, we pose this task as a distribution matching problem such that the synthetic data are optimized to match the original data distribution in a family of embedding spaces by using the maximum mean discrepancy~(MMD)~\cite{gretton2012kernel} measurement (see~\Cref{fig:method}).
Distance between data distributions is commonly used as the criterion for coreset selection~\cite{chen2010super, farahani2009facility, wang2015querying, sener2017active}, however, it has not been used to synthesize training data before.
We show that the family of embedding spaces can be efficiently obtained by sampling randomly initialized deep neural networks.
Hence, our method is significantly faster (\eg 45$\times$ in CIFAR10 when synthesizing 500 images) than the state-of-the-art \cite{zhao2021DSA} and involves tuning only one hyperparameter (learning rate for synthetic images), while obtaining comparable or better results.
In addition, the training of our method can be independently run for each class in parallel and its computation load can be distributed.
Finally, our method provides a different training cost/performance tradeoff for large-scale settings. In contrast to prior works \cite{wang2018dataset, nguyen2021dataset} that are limited to learning small synthetic sets on small datasets, our method can be successfully applied in more realistic settings, \ie synthesizing 1250 images per class for CIFAR10 \cite{krizhevsky2009learning}, and larger datasets, \ie TinyImageNet \cite{le2015tiny} and ImageNet-1K \cite{deng2009imagenet}.
We also validate these benefits in two downstream tasks by producing more data-efficient memory for continual learning and generating more representative proxy dataset for accelerating neural architecture search.

\section{Methodology}
\subsection{Dataset Condensation Problem}
The goal of dataset condensation is to condense the large-scale training set $\mt=\{(\bx_1,y_1),\dots,(\bx_{|\mt|},y_{|\mt|})\}$ with $|\mt|$ image and label pairs into a small synthetic set $\ms=\{(\bs_1,y_1),\dots,(\bs_{|\ms|},y_{|\ms|})\}$ with $|\ms|$ synthetic image and label pairs so that models trained on each $\mt$ and $\ms$ obtain comparable performance on unseen testing data:
\begin{equation}
    \mathbb{E}_{\bx \sim P_{\mathcal{D}}}[\ell(\phi_{\btheta^\mt}(\bx),y)]\simeq\mathbb{E}_{\bx \sim P_{\mathcal{D}}}[\ell(\phi_{\btheta^\ms}(\bx),y)], 
    \label{eq:goalDC}
\end{equation}
where $P_{\mathcal{D}}$ is the real data distribution, $\ell$ is the loss function (\ie cross-entropy loss), $\phi$ is a deep neural network parameterized by $\btheta$, and $\phi_{\btheta^\mt}$ and $\phi_{\btheta^\ms}$ are the networks that are trained on $\mt$ and $\ms$ respectively. 

\paragraph{Existing solutions.} Previous works \cite{wang2018dataset, sucholutsky2019soft, such2020generative, bohdal2020flexible, nguyen2021dataset, nguyen2021dataset2} formulate the dataset condensation as a learning-to-learn problem, pose the network parameters $\btheta^\ms$ as a function of synthetic data $\ms$ and  obtain a solution for $\ms$ by minimizing the training loss $\ml^{\mt}$ over the original data $\mt$:
\begin{equation}
\begin{split}
\ms^\ast=\argmin_{\mathcal{S}}\ml^{\mt}(\btheta^\ms (\ms)) 
\\
\quad \text{subject to} \quad
\btheta^\ms(\ms) = \argmin_{\btheta}\ml^\ms(\btheta).
\end{split}
\label{eq.wang}
\end{equation} 
Recently the authors of \cite{zhao2021DC, zhao2021DSA} show that a similar goal can be achieved by matching gradients of the losses over the synthetic and real training data respectively w.r.t. the network parameters $\btheta$, while optimizing $\btheta$ and the synthetic data $\ms$ in an alternating way:
\begin{equation}
\begin{split}
\ms^\ast=\argmin_{\ms} \mathrm{E}_{\btheta_0\sim P_{\btheta_0}}\Big[\sum_{t=0}^{T-1} D(\nabla_{\btheta}\ml^\ms(\btheta_{t}),\nabla_{\btheta}\ml^\mt(\btheta_{t}))\Big] \\
 \text{subject to} \quad
\btheta_{t+1} \leftarrow \texttt{opt-alg}_{\btheta}(\ml^{\ms}(\btheta_{t}),\varsigma_{\btheta},\eta_{\btheta}),
\end{split}
\label{eq.optgrad}
\end{equation}
where $P_{\btheta_0}$ is the distribution of parameter initialization, $T$ is the outer-loop iteration for updating synthetic data, $\varsigma_{\btheta}$ is the inner-loop iteration for updating network parameters, $\eta_{\btheta}$ is the parameter learning rate and $D(\cdot, \cdot)$ measures the gradient matching error. 
Note that all the training algorithms \cite{wang2018dataset, zhao2021DC, zhao2021DSA} have another loop of sampling $\btheta_0$ over the bi-level optimization.

\paragraph{Dilemma.} 
The learning problems in \cref{eq.wang} and \cref{eq.optgrad} involve solving an expensive bi-level optimization: first optimizing the model $\btheta^\ms$ in \cref{eq.wang} or $\btheta_{t}$ in \cref{eq.optgrad} at the inner loop, then optimizing the synthetic data $\ms$ along with additional second-order derivative computation at the outer loop. 
For example, training 50 images/class synthetic set $\ms$ by using the method in \cite{zhao2021DC} requires 500K epochs of updating network parameters $\btheta_{t}$ on $\ms$, in addition to the 50K updating of $\ms$.
Furthermore, Zhao \emph{et al.} \cite{zhao2021DC} need to tune the hyper-parameters of the outer and inner loop optimization (\ie how many steps to update $\ms$ and $\btheta_{t}$) for different learning settings,
which requires cross-validating them and hence multiplies the cost for training synthetic images. 


\subsection{Dataset Condensation with Distribution Matching}
Our goal is to synthesize data that can accurately approximate the distribution of the real training data in a similar spirit to coreset techniques (\eg \cite{welling2009herding,sener2017active}).
However, to this end, we do not limit our method to select a subset of the training samples but to synthesize them as in \cite{wang2018dataset,zhao2021DC}.
As the training images are typically very high dimensional, estimating the real data distribution $P_{\mathcal{D}}$ can be expensive and inaccurate.
Instead, we assume that each training image $\bx \in \Re^d$ can be embedded into a lower dimensional space by using a family of parametric functions $\psi_{\bvartheta}:\Re^d\rightarrow \Re^{d'}$ where $d' \ll d$ and $\bvartheta$ is the parameter.
In other words, each embedding function $\psi$ can be seen as providing a partial interpretation of its input, while their combination provides a complete one.


Now we can estimate the distance between the real and synthetic data distribution with commonly used maximum mean discrepancy~(MMD)~\cite{gretton2012kernel}:
\begin{equation}
    \sup_{\|\psi_{\bvartheta}\|_{\mathcal{H}}\leq 1}(\mathbb{E}[\psi_{\bvartheta}(\mt)] - \mathbb{E}[\psi_{\bvartheta}(\ms)]),
\end{equation} where $\mathcal{H}$ is reproducing kernel Hilbert space.
As we do not have access to ground-truth data distributions, we use the empirical estimate of the MMD:
\begin{equation}
    \mathbb{E}_{\bvartheta \sim P_{\bvartheta}}\|\frac{1}{|\mt|}\sum_{i=1}^{|\mt|}\psi_{\bvartheta}(\bx_i) - \frac{1}{|\ms|}\sum_{j=1}^{|\ms|}\psi_{\bvartheta}(\bs_j)\|^2,
\label{eq:discrepancy}
\end{equation} where $P_{\bvartheta}$ is the distribution of network parameters.

Following \cite{zhao2021DSA}, we also apply the differentiable Siamese augmentation $\ma(\cdot, \omega)$ to real and synthetic data that implements the same randomly sampled augmentation to the real and synthetic minibatch in training, where $\omega \sim \Omega$ is the augmentation parameter such as the rotation degree. 
Thus, the learned synthetic data can benefit from semantic-preserving transformations (\eg cropping) and learn prior knowledge about spatial configuration of samples while training deep neural networks with data augmentation. 
Finally, we solve the following optimization problem:
\begin{equation}
\small	
\begin{split}
 \min_{\ms}\mathbb{E}_{\substack{\bvartheta \sim P_{\bvartheta} \\ \omega \sim \Omega}}\|\frac{1}{|\mt|}\sum_{i=1}^{|\mt|}\psi_{\bvartheta}(\ma(\bx_i, \omega)) 
  - \frac{1}{|\ms|}\sum_{j=1}^{|\ms|}\psi_{\bvartheta}(\ma(\bs_j, \omega))\|^2.
\end{split}
\label{eq:loss}
\end{equation}
We learn the synthetic data $\mathcal{S}$ by minimizing the discrepancy between two distributions in various embedding spaces by sampling $\bvartheta$.
Importantly \cref{eq:loss} can be efficiently solved, as it requires only optimizing $\ms$ but no model parameters and thus avoids expensive bi-level optimization.
This is in contrast to the existing formulations (see \cref{eq.wang} and \cref{eq.optgrad}) that involve bi-level optimizations over network parameters $\btheta$ and the synthetic data $\ms$.

Note that, as we target at image classification problems, we minimize the discrepancy between the real and synthetic samples of the same class only.
We assume that each real training sample is labelled and we also set a label to each synthetic sample and keep it fixed during training.

\subsection{Training Algorithm}
We depict the mini-batch based training algorithm in Algorithm~\ref{algo}. We train the synthetic data for $K$ iterations. In each iteration, we randomly sample the model $\psi_{\bvartheta}$ with parameter $\bvartheta \sim P_{\bvartheta}$. Then, we sample a pair of real and synthetic data batches ($B^\mt_c\sim\mt$ and $B^\ms_c\sim\ms$) and augmentation parameter $\omega_c\sim \Omega$ for every class $c$. The mean discrepancy between the augmented real and synthetic batches of every class is calculated and then summed as loss $\ml$. The synthetic data $\ms$ is updated by minimizing $\ml$ with stochastic gradient descent and learning rate $\eta$.   

\begin{algorithm*}
\KwIn{Training set $\mt$}
\textbf{Required}: Randomly initialized set of synthetic samples $\ms$ for $C$ classes, deep neural network $\psi_{\bvartheta}$ parameterized with $\bvartheta$, probability distribution over parameters $P_{\bvartheta}$, differentiable augmentation $\ma_{\omega}$ parameterized with $\omega$, augmentation parameter distribution $\Omega$, training iterations $K$, learning rate $\eta$.

\For{$k = 0, \cdots, K-1$}
{
	Sample $\bvartheta \sim P_{\bvartheta}$
	
	Sample mini-batch pairs $B^\mt_c\sim\mt$ and $B^\ms_c\sim\ms$ and $\omega_c\sim \Omega$ for every class $c$  

	Compute $\ml=\sum_{c=0}^{C-1}\|\frac{1}{|B^\mt_c|}\sum_{(\bx,y)\in B^\mt_c}\psi_{\bvartheta}(\ma_{\omega_c}(\bx)) - \frac{1}{|B^\ms_c|}\sum_{(\bs,y)\in B^\ms_c}\psi_{\bvartheta}(\ma_{\omega_c}(\bs))\|^2$

    Update $\ms\leftarrow \ms - \eta\nabla_{\ms}\ml$ 
}

\KwOut{$\ms$}
\caption{Dataset condensation with distribution matching}
\label{algo}
\end{algorithm*}

\subsection{Discussion}
\paragraph{Randomly Initialized Networks.} The family of embedding functions $\psi_{\bvartheta}$ can be designed in different ways.
Here we use a deep neural network with different random initializations rather than sampling its parameters from a set of pre-trained networks which is more computationally expensive to obtain.
We experimentally validate that our random initialization strategy produces better or comparable results with the more expensive strategy of using pretrained networks in \Cref{sec:different_distribution}. 
However, one may still question why randomly initialized networks provide meaningful embeddings for distribution matching.
Here we list two reasons based on the observations from previous work.
First, randomly initialized networks are reported to produce powerful representations for multiple computer vision tasks \cite{saxe2011random, cao2018review, amid2022learning}.
Second, such random networks are showed to perform a distance-preserving embedding of the data, \ie smaller distances between samples of same class and larger distances across samples of different classes \cite{giryes2016deep}. In addition, the combination of many weak embeddings provides a complete interpretation of the inputs.

\paragraph{Connection to Gradient Matching.} 
While we match the mean features of the real and synthetic image batches, Zhao \emph{et al.} \cite{zhao2021DC} match the mean gradients of network weights over the two batches.
We find that, given a batch of data from the same class, the mean gradient vector w.r.t. each output neuron in the last layer of a network is equivalent to a weighted mean of features where the weights are a function of classification probabilities predicted by the network and proportional to the distance between prediction and ground-truth.
In other words, while our method weighs each feature equally, Zhao \emph{et al.} \cite{zhao2021DC} assign larger weights to samples whose predictions are inaccurate. Note that these weights dynamically vary for different networks and training iterations.
We provide the derivation in the appendix.

\paragraph{Generative Models.} The classic image synthesizing techniques, includes AutoEncoders \cite{kingma2013auto} and Generative Adversarial Networks (GANs) \cite{goodfellow2014generative}, aim to synthesize real-looking images, while our goal is to generate data-efficient training samples. Regularizing the images to look real may limit the data-efficiency. 
Previous work \cite{zhao2021DC} showed that the images synthesized by cGAN \cite{mirza2014conditional} are not better than the randomly selected real images for training networks. 
We further provide the comparison to state-of-the-art VAE and GAN models and GMMN method \cite{li2015generative} in the appendix. 
Although generative models can be trained to produce data-efficient samples with suitable objectives, \eg \cite{wang2018dataset, zhao2021DC} and ours, it is not trivial work to build it and achieve state-of-the-art results \cite{such2020generative}. We leave it as the future work.

\section{Experiments}
\label{sec:exp}

\subsection{Experimental Settings}
\label{sec:experimental_settings}
\paragraph{Datasets.} We evaluate the classification performance of deep networks that are trained on the synthetic images generated by our method.
We conduct experiments on five datasets including MNIST \cite{lecun1998gradient}, CIFAR10, CIFAR100 \cite{krizhevsky2009learning}, TinyImageNet \cite{le2015tiny} and ImageNet-1K \cite{deng2009imagenet}.
MNIST consists of 60K $28\times28$ gray-scale training images of 10 classes. 
CIFAR10 and CIFAR100 contain 50k $32\times32$ training images from 10 and 100 object categories respectively. 
TinyImageNet and ImageNet-1K have 100K training images from 200 categories and 1.3M  training images from 1K categories respectively. We resize these ImageNet images with $64\times64$ resolution. These two datasets are significantly more challenging than MNIST and CIFAR10/100 due to more diverse classes and higher image resolution.


\paragraph{Experimental Settings.} We first learn 1/10/50 image(s) per class synthetic sets for all datasets by using the same ConvNet architecture in \cite{zhao2021DC}. Then, we use the learned synthetic sets to train  randomly initialized ConvNets from scratch and evaluate them on real test data. The default ConvNet includes three repeated convolutional blocks, and each block involves a 128-kernel convolution layer, instance normalization layer \cite{ulyanov2016instance}, ReLU activation function \cite{nair2010rectified} and average pooling. 
Note that four-block ConvNets are used to adjust to the larger input size ($64\times64$) of TinyImageNet and ImageNet-1K images.
In each experiment, we learn one synthetic set and use it to test 20 randomly initialized networks. We repeat each experiment for 5 times and report the mean testing accuracy of the 100 trained networks. We also do cross-architecture experiments 
in \Cref{sec:cross_arc}.
where we learn the synthetic set on one network architecture and use them to train networks with different architectures. 

\paragraph{Hyper-parameters.} 
Like the standard neural network training, dataset condensation also involves tuning a set of hyperparameters.
Our method needs to tune \textbf{only one hyper-parameter}, \ie learning rate for the synthetic images, for learning different sizes of synthetic sets, while existing methods \cite{wang2018dataset, zhao2021DC, nguyen2021dataset, wang2022cafe, cazenavette2022distillation} have to tune more hyper-parameters such as the steps to update synthetic images and network parameters respectively.
We use a fixed learning rate 1 for optimizing synthetic images for all 1/10/50 images/class learning on all datasets. When learning larger synthetic sets such as 100/200/500/1,000 images per class, we use larger learning rate (\ie 10) due to the relatively smaller distribution matching loss. We train synthetic images for 20,000 iterations on MNIST, CIFAR10/100 and 10,000 iterations on TinyImageNet and ImageNet-1K respectively. The mini-batch size for sampling real data is 256. We initialize the synthetic images using randomly sampled real images with corresponding labels. All synthetic images of a class are used to compute the class mean. We use the same augmentation strategy as \cite{zhao2021DSA}.

\subsection{Comparison to the State-of-the-art}
\paragraph{Competitors.} We compare our method to three standard coreset selection methods, namely, Random Selection, Herding \cite{chen2010super, rebuffi2017icarl, castro2018end, belouadah2020scail} and Forgetting \cite{toneva2019empirical}. 
Herding method greedily adds samples into the coreset so that the mean vector is approaching the whole dataset mean. Toneva \emph{et al.} \cite{toneva2019empirical} count how many times a training sample is learned and then forgotten during network training. The samples that are less forgetful can be dropped.
We also compare to four state-of-the-art training set synthesis methods, namely, DD~\cite{wang2018dataset}, LD~\cite{bohdal2020flexible}, DC~\cite{zhao2021DC} and DSA~\cite{zhao2021DSA}. 
Note that we are aware of concurrent works~\cite{lee2022dataset, kim2022dataset, cazenavette2022distillation} that largely improves the existing bilevel optimization based dataset condensation solutions. Unlike them, we contribute the first solution that has neither bi-level optimization nor second-order derivative, and provide a different training cost/performance tradeoff. Compared to them, our method is significantly simpler and faster. Thus, it is able to scale to large settings \ie learning 1250 images per class for CIFAR10 and large datasets \ie ImageNet-1K, while these concurrent works can't.
More detailed comparison and discussion to other methods \cite{such2020generative, nguyen2021dataset, nguyen2021dataset2}, MMD baseline \cite{gretton2012kernel} and generative baselines including DC-VAE \cite{parmar2021dual}, BigGAN \cite{brock2018large} and GMMN \cite{li2015generative} can be found in the appendix.

\begin{table*}[t]
\renewcommand\arraystretch{0.9}
\centering
\scriptsize
\setlength{\tabcolsep}{2pt}
\begin{tabular}{ccc|ccc|ccccc|c}
\toprule
\multirow{3}{*}{}           & \multirow{2}{*}{Img/Cls} & \multirow{2}{*}{Ratio \%} & \multicolumn{3}{c|}{Coreset Selection}   & \multicolumn{5}{c|}{Training Set Synthesis} & \multirow{2}{*}{Whole Dataset} \\ 
                            & &                & Random        & Herding       & Forgetting     & DD$^\dagger$  & LD$^\dagger$ & DC             & DSA 					& \emph{DM}         &       \\ \midrule
\multirow{3}{*}{MNIST}          & 1   & 0.017  & 64.9$\pm$3.5  & 89.2$\pm$1.6  & 35.5$\pm$5.6   &               & 60.9$\pm$3.2  & \bf{91.7$\pm$0.5}  & 88.7$\pm$0.6  	& 89.7$\pm$0.6  & \multirow{3}{*}{99.6$\pm$0.0} \\
                                & 10  & 0.17   & 95.1$\pm$0.9  & 93.7$\pm$0.3  & 68.1$\pm$3.3   & 79.5$\pm$8.1  & 87.3$\pm$0.7  & 97.4$\pm$0.2  & \bf{97.8$\pm$0.1}  	& 97.5$\pm$0.1  & \\
                                & 50  & 0.83   & 97.9$\pm$0.2  & 94.8$\pm$0.2  & 88.2$\pm$1.2   & -             & 93.3$\pm$0.3  & 98.8$\pm$0.2  & \bf{99.2$\pm$0.1}  	& 98.6$\pm$0.1  &  \\ \midrule 



\multirow{3}{*}{CIFAR10}        & 1   & 0.02   & 14.4$\pm$2.0  & 21.5$\pm$1.2  & 13.5$\pm$1.2   & -             & 25.7$\pm$0.7  & \bf{28.3$\pm$0.5} & \bf{28.8$\pm$0.7} & 26.0$\pm$0.8  & \multirow{3}{*}{84.8$\pm$0.1} \\
                                & 10  & 0.2    & 26.0$\pm$1.2  & 31.6$\pm$0.7  & 23.3$\pm$1.0   & 36.8$\pm$1.2  & 38.3$\pm$0.4  & 44.9$\pm$0.5  & \bf{52.1$\pm$0.5}  	& 48.9$\pm$0.6  &           \\  
                                & 50  & 1      & 43.4$\pm$1.0  & 40.4$\pm$0.6  & 23.3$\pm$1.1   & -             & 42.5$\pm$0.4  & 53.9$\pm$0.5  & 60.6$\pm$0.5          & \bf{63.0$\pm$0.4}  &  \\ \midrule
                                
\multirow{3}{*}{CIFAR100}     & 1   & 0.2    &  4.2$\pm$0.3  &  8.4$\pm$0.3  &  4.5$\pm$0.2   & -             & 11.5$\pm$0.4  & 12.8$\pm$0.3  & \bf{13.9$\pm$0.3}    		& 11.4$\pm$0.3  &    \multirow{3}{*}{56.2$\pm$0.3}\\ 
                              & 10  & 2      & 14.6$\pm$0.5  & 17.3$\pm$0.3  & 15.1$\pm$0.3   & -             & -             & 25.2$\pm$0.3  & \bf{32.3$\pm$0.3}    		& 29.7$\pm$0.3  &                \\  
                              & 50  & 10     & 30.0$\pm$0.4  & 33.7$\pm$0.5  & 30.5$\pm$0.3   & -             & -             & -             &   42.8$\pm$0.4                    & \bf{43.6$\pm$0.4}  &                \\  \midrule

\multirow{3}{*}{TinyImageNet} & 1   & 0.2    &  1.4$\pm$0.1  &  2.8$\pm$0.2  &  1.6$\pm$0.1   & -             & -               & -             & -    		            &  \bf{3.9$\pm$0.2}  &    \multirow{3}{*}{37.6$\pm$0.4}\\ 
                              & 10  & 2      &  5.0$\pm$0.2  &  6.3$\pm$0.2  &  5.1$\pm$0.2   & -             & -               & -             & -   		            & \bf{12.9$\pm$0.4}  &                \\  
                              & 50  & 10     & 15.0$\pm$0.4  & 16.7$\pm$0.3  & 15.0$\pm$0.3   & -             & -               & -             &   -                   & \bf{24.1$\pm$0.3}  &                \\  \bottomrule


\end{tabular}
\caption{{Comparing to coreset selection and training set synthesis methods. We first learn the synthetic data and then evaluate them by training neural networks from scratch and testing on real testing data. The testing accuracies (\%) are reported. Img/Cls: image(s) per class. Ratio~(\%): the ratio of condensed set size to the whole training set size. Note: DD$^\dagger$ and LD$^\dagger$ use different architectures \ie LeNet for MNIST and AlexNet for CIFAR10. The rest methods all use ConvNet. }}
\label{tab:sota}
\vspace{-5pt}
\end{table*}

\paragraph{Performance Comparison.}
Here we evaluate our method on MNIST, CIFAR10 and CIFAR100 datasets and report the results in \Cref{tab:sota}. Among the coreset selection methods, Herding performances the best in most settings. Especially, when small synthetic sets are learned, Herding method performs significantly better. For example, Herding achieves 8.4\% testing accuracy when learning 1 image/class synthetic set on CIFAR100, while Random and Forgetting obtains only 4.2\% and 4.5\% testing accuracies respectively. 

Training set synthesis methods have clear superiority over coreset selection methods, as the synthetic training data are not limited to a set of real images. 
Best results are obtained either by DSA or our method.
While DSA produces more data-efficient samples with a small number of synthetic samples (1/10 image(s) per class), our method outperforms DSA at 50 images/class setting in CIFAR10 and CIFAR100.
The possible reason is that the inner-loop model optimization in DSA with limited number of steps is more effective to fit the network parameters on smaller synthetic data (see~\cref{eq.optgrad}).
In case of bigger learned synthetic data, the solution obtained in the inner-loop becomes less accurate as it can use only limited number of steps to keep the algorithm scalable.
In contrast, our method is robust to increasing synthetic data size, can be efficiently optimized significantly faster than DSA.


\paragraph{TinyImageNet and ImageNet-1K.} 
Due to higher image resolution and more diverse classes, prior bilevel optimization based methods do not scale to TinyImageNet and ImageNet-1K.
Our method takes 27 hours with one Tesla V100 GPU to condense TinyImageNet into three condensed sets (1/10/50 images/class synthetic sets), and it takes 28 hours with ten GTX 1080 GPUs to condense ImageNet-1K into these three sets.
As shown in \Cref{tab:sota}, our method achieves 3.9\%, 12.9\% and 24.1\% testing accuracies when learning 1, 10 and 50 images/class synthetic sets for TinyImageNet, and recovers 60\% classification performance of the baseline that is trained on the whole original training set with only 10\% of data. Our method significantly outperforms the best coreset selection method - Herding, which obtains 2.8\%, 6.3\% and 16.7\% testing accuracies. On ImageNet-1K dataset, our method achieves 1.3\%, 5.7\% and 11.4\% testing accuracies when learning 1, 10 and 50 images/class synthetic sets, which outperforms random selection (0.52\%, 1.94\% and 7.54\%) by large margins.

\begin{figure}[]
    \vspace{-5pt}
    \centering
    \includegraphics[width=1\linewidth]{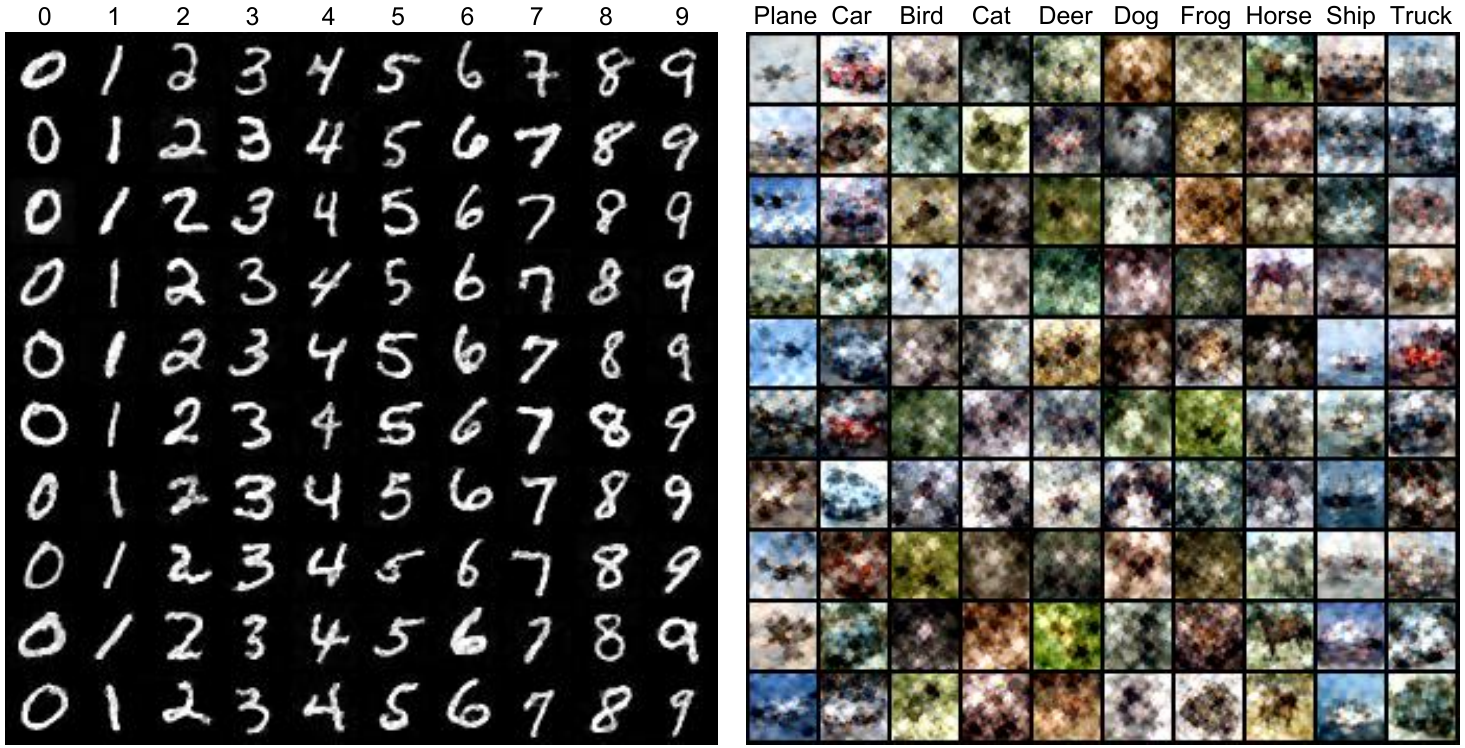}
    \caption{{Visualization of generated 10 images per class synthetic sets of MNIST and CIFAR10 datasets.}}
    \label{fig:vis_sota}
    \vspace{-5pt}
\end{figure}

\paragraph{Visualization.} The learned synthetic images of MNIST and CIFAR10 are visualized in \Cref{fig:vis_sota}. We find that the synthetic MNIST images are clear and noise free, while the number images synthesized by previous methods contain obvious noise and some unnatural strokes. The synthetic images of CIFAR10 dataset are also visually recognizable and diverse. It is easy to distinguish the background and foreground object. 

\begin{figure*}[]
    \centering
    \includegraphics[width=0.7\linewidth]{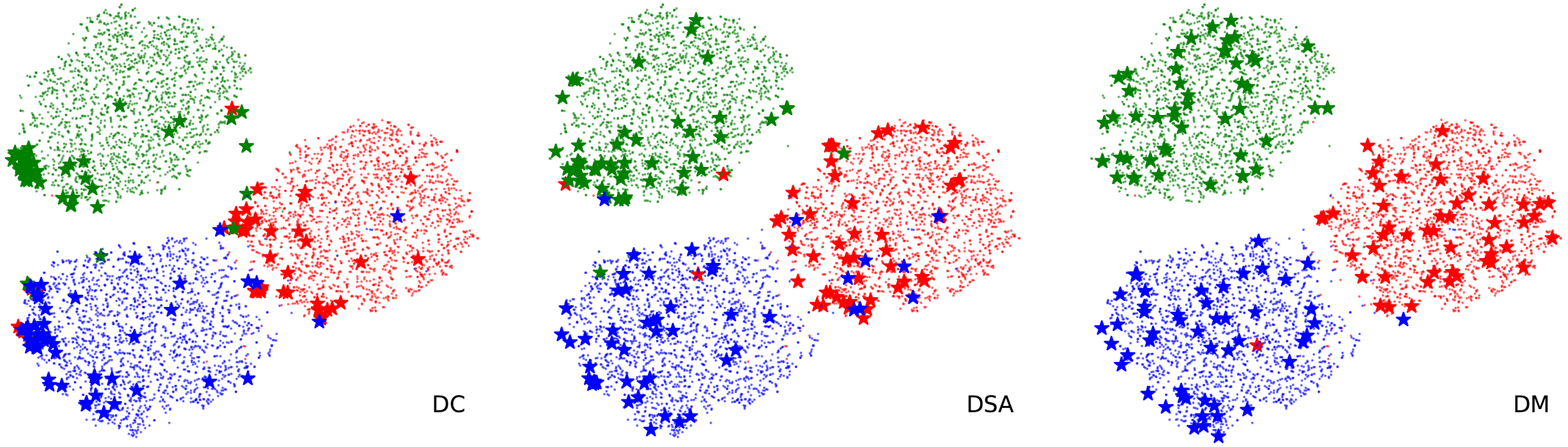}
    \caption{{Distributions of synthetic images learned by DC, DSA and DM. The red, green and blue points are the real images of first three classes in CIFAR10. The stars are corresponding learned synthetic images.}}
    \label{fig:distribution}
    \vspace{-5pt}
\end{figure*}

\Cref{fig:distribution} depicts the feature distribution of the (50 images/class) synthetic sets learned by DC, DSA and our method (DM). We use a network trained on the whole training set to extract features and visualize the features with T-SNE \cite{van2008visualizing}. We find that the synthetic images learned by DC and DSA cannot cover the real image distribution. In contrast, our synthetic images successfully cover the real image distribution. Furthermore, fewer outlier synthetic samples are produced by our method.

\begin{table}[]
\renewcommand\arraystretch{0.7}
\setlength{\tabcolsep}{3pt}
\centering
\scriptsize
\begin{tabular}{c|cc|cc}
\toprule
  & \multicolumn{2}{c|}{InstanceNorm} & \multicolumn{2}{c}{BatchNorm} \\ 
            & DSA            & DM                & DSA           & DM           \\\midrule
CIFAR10        & 60.6$\pm$0.5   & \bf{63.0$\pm$0.4}  & 59.9$\pm$0.8 & \bf{65.2$\pm$0.4}         \\
CIFAR100       & 42.8$\pm$0.4   & \bf{43.6$\pm$0.4}  & 44.6$\pm$0.5 & \bf{48.0$\pm$0.4}       \\
TinyImageNet   & -              & \bf{24.1$\pm$0.3}  & -            & \bf{28.2$\pm$0.5} \\
\bottomrule
\end{tabular}
\caption{{50 images/class learning with Batch Normalization.}}
\label{tab:bn}
\vspace{-5pt}
\end{table}

\begin{table}[]
\centering
\scriptsize
\setlength{\tabcolsep}{3pt}
\renewcommand\arraystretch{0.7}
\begin{tabular}{cccccc}
\toprule
&   \texttt{C}\textbackslash \texttt{T} & ConvNet      & AlexNet      & VGG          & ResNet            \\ 	\midrule
DSA                 & ConvNet           & 59.9$\pm$0.8 & 53.3$\pm$0.7 & 51.0$\pm$1.1 & 47.3$\pm$1.0  \\\midrule
\multirow{4}{*}{DM} & ConvNet           & 65.2$\pm$0.4 & 61.3$\pm$0.6 & 59.9$\pm$0.8 & 57.0$\pm$0.9  \\
                    & AlexNet           & 60.5$\pm$0.4 & 59.8$\pm$0.6 & 58.9$\pm$0.4 & 54.6$\pm$0.7  \\
                    & VGG               & 54.2$\pm$0.6 & 52.6$\pm$1.0 & 52.8$\pm$1.1 & 49.1$\pm$1.0  \\
                    & ResNet            & 52.2$\pm$1.0 & 50.9$\pm$1.4 & 49.6$\pm$0.9 & 52.2$\pm$0.4  \\ \bottomrule
\end{tabular}
\caption{{Cross-architecture testing performance (\%) on CIFAR10. 
The 50 img/cls synthetic set is learned on one architecture (C), and then tested on another architecture (T). 
}}
\label{tab:crsarc}
\vspace{-5pt}
\end{table}

\paragraph{Learning with Batch Normalization.} Zhao \emph{et al.} \cite{zhao2021DC} showed that instance normalization \cite{ulyanov2016instance} works better than batch normalization (BN) \cite{Ioffe2015BatchNA} when learning small synthetic sets because the synthetic data number is too small to calculate stable running mean and standard deviation (std). 
When learning with batch normalization, they first pre-set the BN mean and std using many real training data and then freeze them for synthetic data. Thus, the inaccurate mean and std will make optimization difficult \cite{Ioffe2015BatchNA}. 
In contrast, we estimate running mean and std by inputting augmented synthetic data from all classes. Hence, our method benefits from the true mean and std of synthetic data. 
\Cref{tab:bn} show that using ConvNet with BN can further improve our performance. Specifically, our method with BN achieves 65.2\%, 48.0\% and 28.2\% testing accuracies when learning 50 images/class synthetic sets on CIFAR10, CIFAR100 and TinyImageNet respectively, which means $2.2\%$, $4.4\%$ and 4.1\% improvements over our method with the default instance normalization, and also outperforms DSA with BN by 5.3\% and 3.4\% on CIFAR10 and CIFAR100 respectively.

\begin{figure}[]
    \vspace{-7pt}
    \centering
    \includegraphics[width=0.65\linewidth]{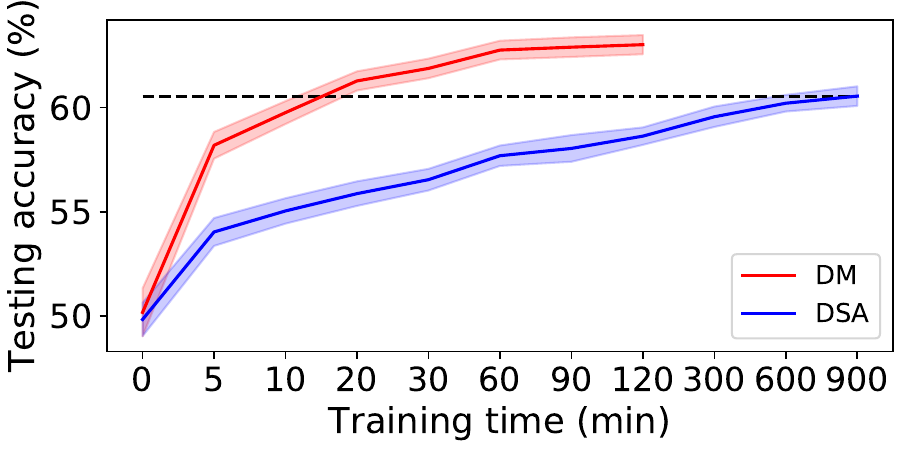}
    \caption{{Training time comparison to DSA when learning 50 img/cls synthetic sets on CIFAR10.}}
    \label{fig:train_time}
    \vspace{-5pt}
\end{figure}

\paragraph{Training Cost Comparison.}
Our method is significantly more efficient than those bi-level optimization based methods. Without loss of generality, we compare the training time of ours and DSA in the setting of learning 50 images/class synthetic data on CIFAR10. \Cref{fig:train_time} shows that our method needs less than 20 minutes to reach the performance of DSA trained for 15 hours, which means less than $2.2\%$ training cost. Note that we run the two methods in the same computation environment with one GTX 1080 GPU.

\thisfloatsetup{floatrowsep=quad} 
\begin{figure*}[]
\centering
    \begin{floatrow}[3] 
	  	\ffigbox[.3\textwidth]{%
		\includegraphics[width=0.3\textwidth]{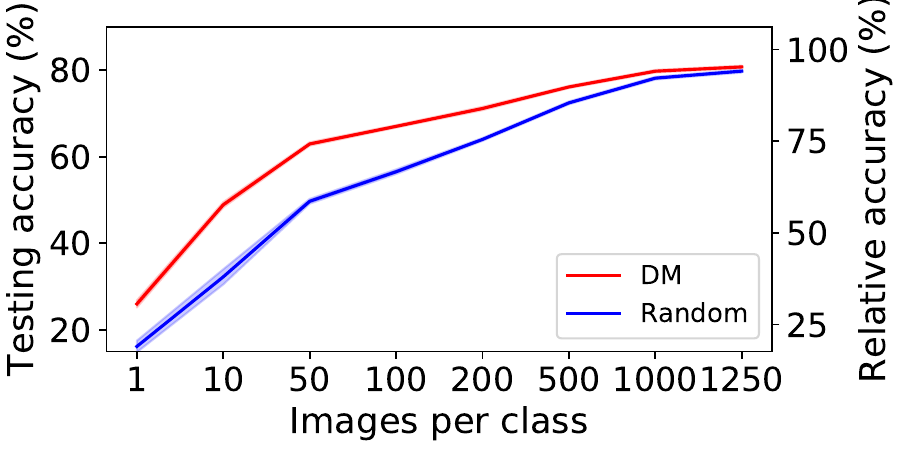}
		}
		{
        \vspace{-20pt}
        \caption{{Learning larger synthetic sets on CIFAR10.}} 
        \label{fig:larger_settings}
		}

	  	\ffigbox[.3\textwidth]{%
		\includegraphics[width=0.3\textwidth]{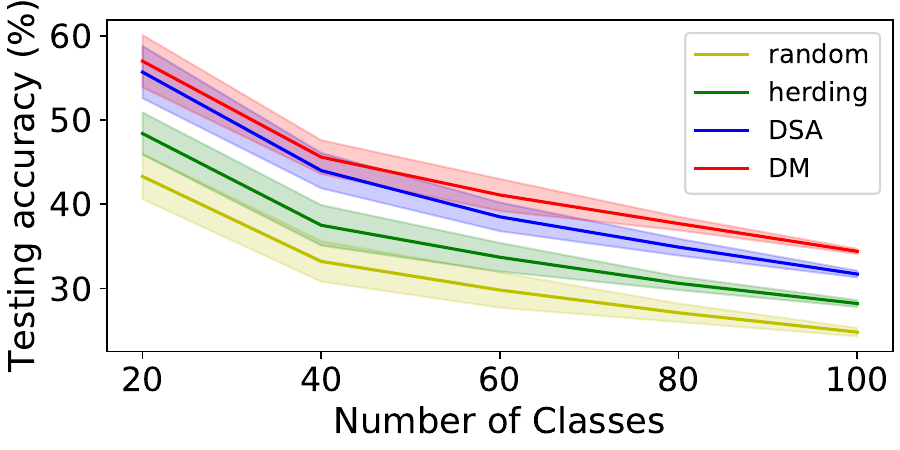}
		}
		{
        \vspace{-20pt}
        \caption{{5-step class-incremental learning on CIFAR100.}}
        \label{fig:cl_5step}
		}

	  	\ffigbox[.3\textwidth]{%
		\includegraphics[width=0.3\textwidth]{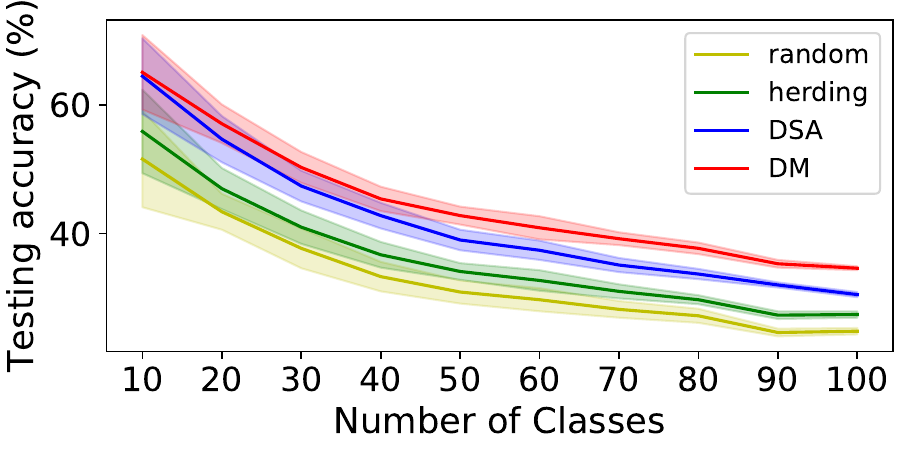}
		}
		{
        \vspace{-20pt}
        \caption{{10-step class-incremental learning on CIFAR100.}}
        \label{fig:cl_10step}
		}
	\end{floatrow}
\end{figure*}

\paragraph{Learning Larger Synthetic Sets}
We show that our method can also be used to learn larger synthetic sets, while the bi-level optimization based methods typically requires more training time and elaborate hyper-parameter tuning for larger settings. 
\Cref{fig:larger_settings} compares our method to random selection baseline in CIFAR10 in terms of absolute and relative performance w.r.t. whole dataset training performance.
Clearly our method outperforms random baseline at all operating points which means that our synthetic set is more data-efficient. The advantage of our method is remarkable in challenging settings, \ie settings with small data budgets.
Our method obtains $67.0\pm0.3\%$, $71.2\pm0.4\%$, $76.1\pm0.3\%$, $79.8\pm0.3\%$ and $80.8\pm0.3\%$ testing accuracies when learning 100, 200, 500, 1000 and 1250 images/class synthetic sets on CIFAR10 dataset respectively, which means we can recover $79\%$, $84\%$, $90\%$, $94\%$ and $95\%$ relative performance using only $2\%$, $4\%$, $10\%$, $20\%$ and $25\%$ training data compared to whole dataset training.
We see that the performance gap between the two methods narrows when we learn larger synthetic set. 
This is somewhat expected, as randomly selecting more samples will approach the whole dataset training which can be considered as the upper-bound. As we initialize synthetic images from random real images, the initial distribution discrepancy becomes tiny when the synthetic set is large. 

\subsection{Cross-architecture Generalization}
\label{sec:cross_arc}
\cite{zhao2021DC, zhao2021DSA} verified the cross-architecture generalization ability of synthetic data in an easy setting - learning 1 image/class for MNIST dataset. In this paper, we implement a more challenging cross-architecture experiment - learning 50 images/class for CIFAR10 dataset.
In \Cref{tab:crsarc}, the synthetic data are learned with one architecture (denoted as C) and then be evaluated on another architecture (denoted as T) by training a model from scratch and testing on real testing data.
We test several sophisticated neural architectures namely ConvNet, AlexNet \cite{krizhevsky2012imagenet}, VGG-11 \cite{simonyan2014very} and ResNet-18 \cite{he2016deep}. Batch Normalization is used in all architectures.

\Cref{tab:crsarc} shows that learning and evaluating synthetic set on ConvNet achieves the best performance 65.2\%. 
Comparing with DSA, the synthetic data learned by our method with ConvNet have better generalization performance than that learned by DSA with the ConvNet. Specifically, our method outperforms DSA by 8.0\%, 8.9\% and 9.7\% when testing with AlexNet, VGG and ResNet respectively. 
These results indicate that the synthetic images learned with distribution matching have better generalization performance on unseen architectures than those learned with gradient matching.
The learning of synthetic set can be worse with more sophisticated architecture such as ResNet. It is reasonable that the synthetic data fitted on sophisticated architecture will contain some bias that doesn't exist in other architectures, therefore cause worse cross-architecture generalization performance. We also find that the evaluation of the same synthetic set on more sophisticated architectures will be worse. The reason may be that sophisticated architectures are under-fitted using small synthetic set.

\subsection{Ablation Study on Network Distribution}
\label{sec:different_distribution}
Here we study the effect of using different network distributions while learning 1/10/50 image(s)/class synthetic sets on CIFAR10 with ConvNet architecture.
Besides sampling randomly initialized network parameters, we also construct a set of networks that are pre-trained on the original training set.
In particular, we train 1,000 ConvNets with different random initializations on the whole original training set and also store their intermediate states.
We roughly divide these networks into nine groups according to their validation accuracies, sample networks from each group, learn the synthetic data on them and use learned synthetic data to train randomly initialized neural networks.
Interestingly we see in \Cref{tab:distribution} that our method works well with all nine network distributions and the performance variance is small. 
The visualization and analysis about synthetic images learned with different network distributions are provided in the appendix.

\begin{table}
\vspace{-7pt}
\renewcommand\arraystretch{0.7}
\centering
\scriptsize
\setlength{\tabcolsep}{3pt}
\begin{tabular}{cccccccccc}
\toprule
   & Random & 10-20 & 20-30 & 30-40 & 40-50 & 50-60 & 60-70 & $\geq$70 & All \\ \midrule
1  & 26.0   & 26.2  & 25.9  & 26.1  & 26.7  & 26.8  & 27.3  & 26.5  & 26.4                              \\
10 & 48.9   & 48.7  & 48.1  & 50.7  & 51.1  & 49.9  & 48.6  & 48.2  & 50.7                              \\
50 & 63.0   & 62.7  & 62.1  & 62.8  & 63.0  & 61.9  & 60.6  & 60.0  & 62.5                             \\\bottomrule
\end{tabular}
\caption{{The performance of synthetic data learned on CIFAR10 with different network distributions. All standard deviations in this table are $<1$. These networks are trained on the whole training set and grouped based on the validation accuracy (\%).}}
\label{tab:distribution}
\vspace{-5pt}
\end{table}


\subsection{Continual Learning}
We also use our method to store more efficient training samples in the memory for relieving the catastrophic forgetting problem in continual (incremental) learning \cite{rebuffi2017icarl}.
We set up the baseline based on GDumb \cite{prabhu2020gdumb} which stores training samples in memory greedily and keeps class-balance. The model is trained from scratch on the latest memory only. Hence, the continual learning performance completely depends on the quality of the memory construction. 
We compare our memory construction method \ie training set condensation to the \emph{random} selection that is used in \cite{prabhu2020gdumb}, \emph{herding} \cite{chen2010super, rebuffi2017icarl, castro2018end, belouadah2020scail} and \emph{DSA} \cite{zhao2021DSA}. 
We implement class-incremental learning on CIFAR100 dataset with an increasing memory budget of 20 images/class. We implement 5 and 10 step learning, in which we randomly and evenly split the 100 classes into 5 and 10 learning steps \ie 20 and 10 classes per step respectively. The default ConvNet is used in this experiment.

As depicted in \Cref{fig:cl_5step} and \Cref{fig:cl_10step}, we find that our method GDumb + DM outperforms others in both two settings, which means that our method can produce the best condensed set as the memory. The final performances of ours, DSA, herding and random are 34.4\%, 31.7\%, 28.2\% and 24.8\% in 5-step learning and 34.6\%, 30.5\%, 27.4\% and 24.8\% in 10-step learning. We find that ours and random selection performances are not influenced by how the classes are split namely how many new training classes and images occur in each learning step, because both two methods learn/generate the sets independently for each class. However, DSA and herding methods perform worse when the training classes are densely split into more learning steps. The reason is that DSA and herding needs to learn/generate sets based on the model(s) trained on the current training data, which is influenced by the data split. More details can be found in the appendix.

\begin{table}[]
\renewcommand\arraystretch{0.7}
\centering
\scriptsize
\setlength{\tabcolsep}{2pt}
\begin{tabular}{ccccc|c}
\toprule
                    & Random            & DSA               & DM            & Early-stopping    & Whole Dataset \\ \midrule
Performance (\%)    & 84.0               & 82.6              & 82.8          & \bf{84.3}         & 85.9  \\
Correlation         & -0.04              & 0.68              & \bf{0.76}     & 0.11              & 1.00 \\
Time cost (min)     & 142.6             & 142.6             & 142.6         & 142.6             & 3580.2 \\ 
Storage (imgs)      & \bf{500}          & \bf{500}          & \bf{500}      & $5\times 10^4$            & $5\times 10^4$   \\ \bottomrule
\end{tabular}
\vspace{-5pt}
\caption{{We implement neural architecture search on CIFAR10 dataset with the search space of 720 ConvNets.}}
\label{tab:nas}
\end{table}

\subsection{Neural Architecture Search}
The synthetic sets can also be used as a proxy set to accelerate model evaluation in Neural Architecture Search (NAS) \cite{elsken2019neural}. Following \cite{zhao2021DC}, we implement NAS on CIFAR10 with the search space of 720 ConvNets varying in network depth, width, activation, normalization and pooling layers. 
Please refer to \cite{zhao2021DC} for more details.
We train all architectures on the learned 50 images/class synthetic set, \ie 1\% size of the whole dataset, from scratch and then rank them based on the accuracy on a small validation set. We compare to \emph{random}, \emph{DSA} and \emph{early-stopping} methods. The same size of real images are selected as the proxy set in \emph{random}. 
\emph{DSA} means that we use the synthetic set learned by DSA in the same setting. 
In \emph{early-stopping}, we use the whole training set to train the model but with the same training iterations like training on the proxy datasets. Therefore, all these methods have the same training time. We train models on the proxy sets for 200 epochs and whole dataset for 100 epochs. The best model is selected based on validation accuracies obtained by different methods. The Spearman’s rank correlation between performances of proxy-set and whole-dataset training is computed for the top 5\% architectures selected by the proxy-set. 

The NAS results are provided in \Cref{tab:nas}. Although the architecture selected by early-stopping achieves the best performance (84.3\%), its performance rank correlation (0.11) is remarkably lower than DSA (0.68) and DM (0.76). In addition, early-stopping needs to use the whole training set, while other proxy-set methods need only 500 training samples. 
The performance rank correlation of Random (-0.04) is too low to provide a reliable ranking for the architectures.
Our method (DM) achieves the highest performance rank correlation (0.76), which means that our method can produce reliable ranking for those candidate architectures while using only around $\frac{1}{25}$ training time of whole dataset training. 
Although our method needs 72 min to obtain the condensed set, it is negligible compared to whole-dataset training (3580.2 min). 
More implementation details and analysis can be found in the appendix.

\section{Conclusion}
In this paper, we propose an efficient dataset condensation method based on distribution matching. To our knowledge, it is the first solution that has neither bi-level optimization nor second-order derivative. Thus, the synthetic data of different classes can be learned independently and in parallel.
Thanks to its efficiency, we can apply our method to more challenging datasets - TinyImageNet and ImageNet-1K, and learn larger synthetic sets - 1250 images/class on CIFAR10.
Our method is ${45}$ times faster than the state-of-the-art for learning 50 images/class synthetic set on CIFAR10. We also empirically prove that our method can produce more informative memory for continual learning and better proxy set for speeding up model evaluation in NAS.  
Though remarkable progress has been seen in this area since the pioneering work \cite{wang2018dataset} released in 2018, dataset condensation is still in its early stage. 
We will extend dataset condensation to more complex vision tasks in the future.

\paragraph{Acknowledgment.} This work is funded by China Scholarship Council 201806010331 and the EPSRC programme grant Visual AI EP/T028572/1.

{\small
\bibliographystyle{ieee_fullname}
\bibliography{refs}

\begin{thebibliography}{10}\itemsep=-1pt

\bibitem{aljundi2019gradient}
Rahaf Aljundi, Min Lin, Baptiste Goujaud, and Yoshua Bengio.
\newblock Gradient based sample selection for online continual learning.
\newblock In {\em Advances in Neural Information Processing Systems}, pages
  11816--11825, 2019.

\bibitem{amid2022learning}
Ehsan Amid, Rohan Anil, Wojciech Kot{\l}owski, and Manfred~K Warmuth.
\newblock Learning from randomly initialized neural network features.
\newblock {\em arXiv preprint arXiv:2202.06438}, 2022.

\bibitem{AIandCompute}
Dario Amodei, Danny Hernandez, Girish Sastry, Jack Clark, Greg Brockman, and
  Ilya Sutskever.
\newblock Ai and compute.
\newblock In {\em OpenAI Blog}, 2018.

\bibitem{belouadah2020scail}
Eden Belouadah and Adrian Popescu.
\newblock Scail: Classifier weights scaling for class incremental learning.
\newblock In {\em The IEEE Winter Conference on Applications of Computer
  Vision}, 2020.

\bibitem{bergstra2012random}
James Bergstra and Yoshua Bengio.
\newblock Random search for hyper-parameter optimization.
\newblock {\em Journal of machine learning research}, 13(Feb):281--305, 2012.

\bibitem{bohdal2020flexible}
Ondrej Bohdal, Yongxin Yang, and Timothy Hospedales.
\newblock Flexible dataset distillation: Learn labels instead of images.
\newblock {\em Neural Information Processing Systems Workshop}, 2020.

\bibitem{borsos2020coresets}
Zal{\'a}n Borsos, Mojmir Mutny, and Andreas Krause.
\newblock Coresets via bilevel optimization for continual learning and
  streaming.
\newblock {\em Advances in Neural Information Processing Systems},
  33:14879--14890, 2020.

\bibitem{brock2018large}
Andrew Brock, Jeff Donahue, and Karen Simonyan.
\newblock Large scale gan training for high fidelity natural image synthesis.
\newblock {\em ICLR}, 2019.

\bibitem{cao2018review}
Weipeng Cao, Xizhao Wang, Zhong Ming, and Jinzhu Gao.
\newblock A review on neural networks with random weights.
\newblock {\em Neurocomputing}, 275:278--287, 2018.

\bibitem{castro2018end}
Francisco~M Castro, Manuel~J Mar{\'\i}n-Jim{\'e}nez, Nicol{\'a}s Guil, Cordelia
  Schmid, and Karteek Alahari.
\newblock End-to-end incremental learning.
\newblock In {\em Proceedings of the European Conference on Computer Vision
  (ECCV)}, pages 233--248, 2018.

\bibitem{cazenavette2022distillation}
George Cazenavette, Tongzhou Wang, Antonio Torralba, Alexei~A. Efros, and
  Jun-Yan Zhu.
\newblock Dataset distillation by matching training trajectories.
\newblock In {\em Proceedings of the IEEE/CVF Conference on Computer Vision and
  Pattern Recognition}, 2022.

\bibitem{chen2010super}
Yutian Chen, Max Welling, and Alex Smola.
\newblock Super-samples from kernel herding.
\newblock {\em The Twenty-Sixth Conference Annual Conference on Uncertainty in
  Artificial Intelligence}, 2010.

\bibitem{deng2009imagenet}
Jia Deng, Wei Dong, Richard Socher, Li-Jia Li, Kai Li, and Li Fei-Fei.
\newblock Imagenet: A large-scale hierarchical image database.
\newblock In {\em Computer Vision and Pattern Recognition, 2009. CVPR 2009.
  IEEE Conference on}, pages 248--255. Ieee, 2009.

\bibitem{elsken2019neural}
Thomas Elsken, Jan~Hendrik Metzen, Frank Hutter, et~al.
\newblock Neural architecture search: A survey.
\newblock {\em J. Mach. Learn. Res.}, 20(55):1--21, 2019.

\bibitem{farahani2009facility}
Reza~Zanjirani Farahani and Masoud Hekmatfar.
\newblock {\em Facility location: concepts, models, algorithms and case
  studies}.
\newblock Springer Science \& Business Media, 2009.

\bibitem{giryes2016deep}
Raja Giryes, Guillermo Sapiro, and Alex~M Bronstein.
\newblock Deep neural networks with random gaussian weights: A universal
  classification strategy?
\newblock {\em IEEE Transactions on Signal Processing}, 64(13):3444--3457,
  2016.

\bibitem{goodfellow2014generative}
Ian Goodfellow, Jean Pouget-Abadie, Mehdi Mirza, Bing Xu, David Warde-Farley,
  Sherjil Ozair, Aaron Courville, and Yoshua Bengio.
\newblock Generative adversarial nets.
\newblock In {\em Advances in neural information processing systems}, pages
  2672--2680, 2014.

\bibitem{gretton2012kernel}
Arthur Gretton, Karsten~M Borgwardt, Malte~J Rasch, Bernhard Sch{\"o}lkopf, and
  Alexander Smola.
\newblock A kernel two-sample test.
\newblock {\em The Journal of Machine Learning Research}, 13(1):723--773, 2012.

\bibitem{he2016deep}
Kaiming He, Xiangyu Zhang, Shaoqing Ren, and Jian Sun.
\newblock Deep residual learning for image recognition.
\newblock In {\em Proceedings of the IEEE conference on computer vision and
  pattern recognition}, pages 770--778, 2016.

\bibitem{Ioffe2015BatchNA}
Sergey Ioffe and Christian Szegedy.
\newblock Batch normalization: Accelerating deep network training by reducing
  internal covariate shift.
\newblock {\em ArXiv}, abs/1502.03167, 2015.

\bibitem{kim2022dataset}
Jang-Hyun Kim, Jinuk Kim, Seong~Joon Oh, Sangdoo Yun, Hwanjun Song, Joonhyun
  Jeong, Jung-Woo Ha, and Hyun~Oh Song.
\newblock Dataset condensation via efficient synthetic-data parameterization.
\newblock In {\em Proceedings of the International Conference on Machine
  Learning (ICML)}, pages 11102--11118, 2022.

\bibitem{kingma2013auto}
Diederik~P Kingma and Max Welling.
\newblock Auto-encoding variational bayes.
\newblock {\em arXiv preprint arXiv:1312.6114}, 2013.

\bibitem{krizhevsky2009learning}
Alex Krizhevsky, Geoffrey Hinton, et~al.
\newblock Learning multiple layers of features from tiny images.
\newblock Technical report, Citeseer, 2009.

\bibitem{krizhevsky2012imagenet}
Alex Krizhevsky, Ilya Sutskever, and Geoffrey~E Hinton.
\newblock Imagenet classification with deep convolutional neural networks.
\newblock In {\em Advances in neural information processing systems}, pages
  1097--1105, 2012.

\bibitem{le2015tiny}
Ya Le and Xuan Yang.
\newblock Tiny imagenet visual recognition challenge.
\newblock {\em CS 231N}, 7(7):3, 2015.

\bibitem{lecun1998gradient}
Yann LeCun, L{\'e}on Bottou, Yoshua Bengio, Patrick Haffner, et~al.
\newblock Gradient-based learning applied to document recognition.
\newblock {\em Proceedings of the IEEE}, 86(11):2278--2324, 1998.

\bibitem{lee2022dataset}
Saehyung Lee, Sanghyuk Chun, Sangwon Jung, Sangdoo Yun, and Sungroh Yoon.
\newblock Dataset condensation with contrastive signals.
\newblock In {\em Proceedings of the International Conference on Machine
  Learning (ICML)}, pages 12352--12364, 2022.

\bibitem{li2015generative}
Yujia Li, Kevin Swersky, and Rich Zemel.
\newblock Generative moment matching networks.
\newblock In {\em International conference on machine learning}, pages
  1718--1727. PMLR, 2015.

\bibitem{lyu2020threats}
Lingjuan Lyu, Han Yu, and Qiang Yang.
\newblock Threats to federated learning: A survey.
\newblock {\em FL-IJCAI}, 2020.

\bibitem{mirza2014conditional}
Mehdi Mirza and Simon Osindero.
\newblock Conditional generative adversarial nets.
\newblock {\em arXiv preprint arXiv:1411.1784}, 2014.

\bibitem{nair2010rectified}
Vinod Nair and Geoffrey~E Hinton.
\newblock Rectified linear units improve restricted boltzmann machines.
\newblock In {\em Proceedings of the 27th international conference on machine
  learning (ICML-10)}, pages 807--814, 2010.

\bibitem{nguyen2021dataset}
Timothy Nguyen, Zhourong Chen, and Jaehoon Lee.
\newblock Dataset meta-learning from kernel-ridge regression.
\newblock In {\em International Conference on Learning Representations}, 2021.

\bibitem{nguyen2021dataset2}
Timothy Nguyen, Roman Novak, Lechao Xiao, and Jaehoon Lee.
\newblock Dataset distillation with infinitely wide convolutional networks.
\newblock {\em arXiv preprint arXiv:2107.13034}, 2021.

\bibitem{parmar2021dual}
Gaurav Parmar, Dacheng Li, Kwonjoon Lee, and Zhuowen Tu.
\newblock Dual contradistinctive generative autoencoder.
\newblock In {\em Proceedings of the IEEE/CVF Conference on Computer Vision and
  Pattern Recognition}, pages 823--832, 2021.

\bibitem{prabhu2020gdumb}
Ameya Prabhu, Philip~HS Torr, and Puneet~K Dokania.
\newblock Gdumb: A simple approach that questions our progress in continual
  learning.
\newblock In {\em European Conference on Computer Vision}, pages 524--540.
  Springer, 2020.

\bibitem{rebuffi2017icarl}
Sylvestre-Alvise Rebuffi, Alexander Kolesnikov, Georg Sperl, and Christoph~H
  Lampert.
\newblock icarl: Incremental classifier and representation learning.
\newblock In {\em Proceedings of the IEEE Conference on Computer Vision and
  Pattern Recognition}, pages 2001--2010, 2017.

\bibitem{saxe2011random}
Andrew~M Saxe, Pang~Wei Koh, Zhenghao Chen, Maneesh Bhand, Bipin Suresh, and
  Andrew~Y Ng.
\newblock On random weights and unsupervised feature learning.
\newblock In {\em Icml}, 2011.

\bibitem{sener2017active}
Ozan Sener and Silvio Savarese.
\newblock Active learning for convolutional neural networks: A core-set
  approach.
\newblock {\em ICLR}, 2018.

\bibitem{simonyan2014very}
Karen Simonyan and Andrew Zisserman.
\newblock Very deep convolutional networks for large-scale image recognition.
\newblock {\em arXiv preprint arXiv:1409.1556}, 2014.

\bibitem{such2020generative}
Felipe~Petroski Such, Aditya Rawal, Joel Lehman, Kenneth~O Stanley, and Jeff
  Clune.
\newblock Generative teaching networks: Accelerating neural architecture search
  by learning to generate synthetic training data.
\newblock {\em International Conference on Machine Learning (ICML)}, 2020.

\bibitem{sucholutsky2019soft}
Ilia Sucholutsky and Matthias Schonlau.
\newblock Soft-label dataset distillation and text dataset distillation.
\newblock {\em arXiv preprint arXiv:1910.02551}, 2019.

\bibitem{toneva2019empirical}
Mariya Toneva, Alessandro Sordoni, Remi Tachet~des Combes, Adam Trischler,
  Yoshua Bengio, and Geoffrey~J Gordon.
\newblock An empirical study of example forgetting during deep neural network
  learning.
\newblock {\em ICLR}, 2019.

\bibitem{ulyanov2016instance}
Dmitry Ulyanov, Andrea Vedaldi, and Victor Lempitsky.
\newblock Instance normalization: The missing ingredient for fast stylization.
\newblock {\em arXiv preprint arXiv:1607.08022}, 2016.

\bibitem{van2008visualizing}
Laurens Van~der Maaten and Geoffrey Hinton.
\newblock Visualizing data using t-sne.
\newblock {\em Journal of machine learning research}, 9(11), 2008.

\bibitem{wang2022cafe}
Kai Wang, Bo Zhao, Xiangyu Peng, Zheng Zhu, Shuo Yang, Shuo Wang, Guan Huang,
  Hakan Bilen, Xinchao Wang, and Yang You.
\newblock Cafe: Learning to condense dataset by aligning features.
\newblock {\em CVPR}, 2022.

\bibitem{wang2018dataset}
Tongzhou Wang, Jun-Yan Zhu, Antonio Torralba, and Alexei~A Efros.
\newblock Dataset distillation.
\newblock {\em arXiv preprint arXiv:1811.10959}, 2018.

\bibitem{wang2015querying}
Zheng Wang and Jieping Ye.
\newblock Querying discriminative and representative samples for batch mode
  active learning.
\newblock {\em ACM Transactions on Knowledge Discovery from Data (TKDD)},
  9(3):1--23, 2015.

\bibitem{welling2009herding}
Max Welling.
\newblock Herding dynamical weights to learn.
\newblock In {\em Proceedings of the 26th Annual International Conference on
  Machine Learning}, pages 1121--1128. ACM, 2009.

\bibitem{ying2019bench}
Chris Ying, Aaron Klein, Eric Christiansen, Esteban Real, Kevin Murphy, and
  Frank Hutter.
\newblock Nas-bench-101: Towards reproducible neural architecture search.
\newblock In {\em International Conference on Machine Learning}, pages
  7105--7114. PMLR, 2019.

\bibitem{yun2019cutmix}
Sangdoo Yun, Dongyoon Han, Seong~Joon Oh, Sanghyuk Chun, Junsuk Choe, and
  Youngjoon Yoo.
\newblock Cutmix: Regularization strategy to train strong classifiers with
  localizable features.
\newblock In {\em Proceedings of the IEEE/CVF International Conference on
  Computer Vision}, pages 6023--6032, 2019.

\bibitem{zhao2021DSA}
Bo Zhao and Hakan Bilen.
\newblock Dataset condensation with differentiable siamese augmentation.
\newblock In {\em International Conference on Machine Learning}, 2021.

\bibitem{zhao2021DC}
Bo Zhao, Konda~Reddy Mopuri, and Hakan Bilen.
\newblock Dataset condensation with gradient matching.
\newblock In {\em International Conference on Learning Representations}, 2021.

\bibitem{zhao2020differentiable}
Shengyu Zhao, Zhijian Liu, Ji Lin, Jun-Yan Zhu, and Song Han.
\newblock Differentiable augmentation for data-efficient gan training.
\newblock {\em Neural Information Processing Systems}, 2020.

\end{thebibliography}
}

\newpage
\appendix

\renewcommand{\thetable}{T\arabic{table}}
\renewcommand{\thefigure}{F\arabic{figure}}

\section{Implementation details}
\label{sec:appendix_imple_details}
\subsection{Dataset Condensation}
\paragraph{DSA Results.}
As \cite{zhao2021DSA} didn't report 50 images/class learning performance on CIFAR100, we obtain the result in Table 1 by running their released code and coarsely searching the hyper-parameters (outer and inner loop steps). Then, we set both outer and inner loop to be 10 steps. The rest hyper-parameters are the default ones in their released code.
To obtain the DSA results with batch normalization in Table 2 and Table 3, we also run DSA code and set batch normalization in ConvNet.

\paragraph{ResNet with Batch Normalization.} We follow the modification of ResNet in \cite{zhao2021DC}. They replace the $stride=2$ convolution layer with $stride=1$ convolution layer followed by an average pooling layer in the ResNet architecture that is used to learn the synthetic data. This modification enables smooth error back-propagation to the input images. We directly use their released ResNet architecture.

\subsection{Continual Learning}
\paragraph{Data Augmentation.} Prabhu \etal \cite{prabhu2020gdumb} use cutmix \cite{yun2019cutmix} augmentation strategy for training models. Different from them, we follow \cite{zhao2021DSA} and use the default DSA augmentation strategy in order to be consistent with other experiments in this paper.

\paragraph{DSA and Herding Training.} Without loss of generality, we run DSA training algorithm on the new training classes and images only in every learning step. It is not a easy work to take old model and memory into DSA training and achieve better performance. The synthetic data learned with old model can also be biased to it, and thus perform worse. Similarly, we train the embedding function (ConvNet) for herding method on the new training classes and images only.

\subsection{Neural Architecture Search}
We randomly select 10\% training samples in CIFAR10 dataset as the validation set. The rest are the training set. The batch size is 250, then one training epoch on the small (50 images/class) proxy sets includes 2 batches. The DSA augmentation strategy is applied to all proxy-set methods and early-stopping. We train each model 5 times and report the mean accuracies. We do NAS experiment on one Tesla v100 GPU. 

\begin{figure*}[h]
    \centering
    \includegraphics[width=0.8\linewidth]{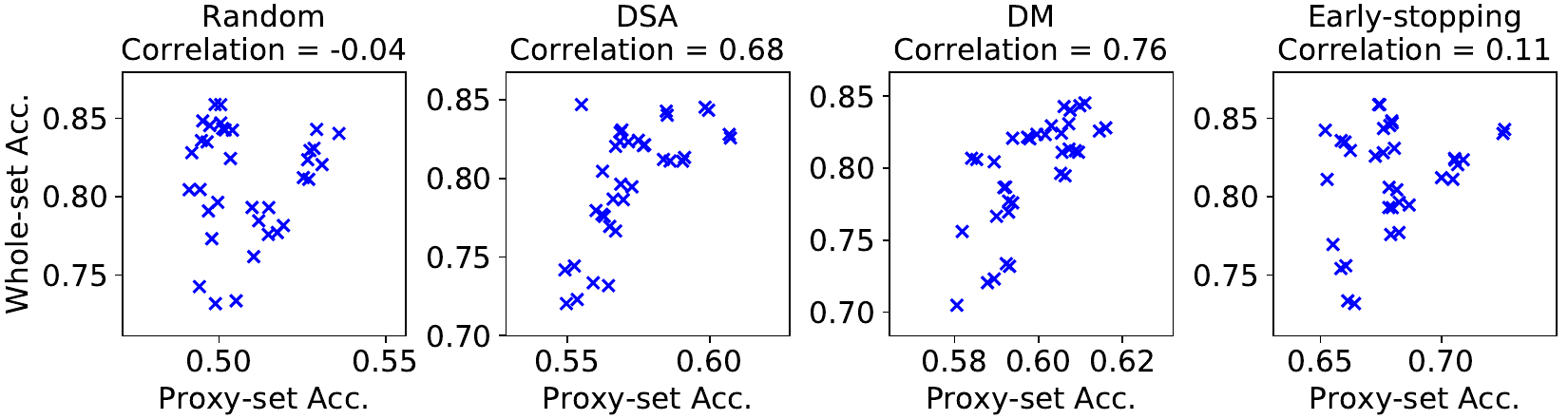}
    \caption{{Performance rank correlation between proxy-set and whole-dataset training.}}
    \label{fig:nas}
\end{figure*}

We visualize the performance rank correlation between proxy-set and whole-dataset training in \Cref{fig:nas}. The top 5\% architectures are selected based on the validation accuracies of models trained on each proxy-set. Each point represents a selected architecture. The horizontal and vertical axes are the testing accuracies of models trained on the proxy-set and the whole dataset respectively. The figure shows that our method can produce better proxy set to obtain more reliable performance ranking of candidate architectures.

\section{Comparison to More Baselines and Related Works}

\subsection{Comparison to Generative Models}
In this subsection, we compare the data-efficiency of samples generated by our dataset condensation method to those generated by traditional generative models, namely VAE and GAN. Specifically, we choose the state-of-the-art DC-VAE  \cite{parmar2021dual} and BigGAN \cite{brock2018large}. The BigGAN model is trained with the differentiable augmentation \cite{zhao2020differentiable}. 
In addition, we also compare to a related generative model GMMN \cite{li2015generative} which aims to learn an image generator that can map a uniform distribution to real image distribution. 
Our method differs from GMMN in many ways significantly.
First, GMMN aims to generate real-looking images, while our goal is to condense a training set by synthesizing informative training samples that can be used to efficiently train deep networks through MMD.
Second, our method learns pixels directly, while GMMN learns a generator network. 
Third, our method learns \textbf{a few} synthetic samples to approximate the distribution of large real training set in \textbf{any} embedding space with \textbf{any} augmentation, while GMMN learns to map a uniform distribution to real image distribution which is an easier task.

We train these generative models on CIFAR10 dataset. ConvNets are trained on these synthetic images and then evaluated on real testing images.
The results in \Cref{tab:vae_gan} verify that our method outperforms them by large margins, indicating that our synthetic images are more informative for training deep neural networks. The comparison to random baseline indicates that the images generated by traditional generative models are not more informative than randomly selected real images.

\begin{table}[h]
\scriptsize
\setlength{\tabcolsep}{3pt}
\renewcommand\arraystretch{0.9}
\begin{tabular}{ccccccc}
\toprule
Img/Cls & Random        & GMMN          & VAE           & BigGAN        & MMD           & \emph{DM} \\ \midrule
1       & 14.4$\pm$2.0  & 16.1$\pm$2.0  & 15.7$\pm$2.1  & 15.8$\pm$1.2  & 22.7$\pm$0.6  & \bf{26.0$\pm$0.8}     \\
10      & 26.0$\pm$1.2  & 32.2$\pm$1.3  & 29.8$\pm$1.0  & 31.0$\pm$1.4  & 34.9$\pm$0.3  & \bf{48.9$\pm$0.6}     \\
50      & 43.4$\pm$1.0  & 45.3$\pm$1.0  & 44.0$\pm$0.8  & 46.2$\pm$0.9  & 50.9$\pm$0.3  & \bf{63.0$\pm$0.4}     \\ \bottomrule
\end{tabular}
\caption{Comparison to traditional generative models and MMD baseline. Random means randomly selected real images. The experiments are implemented with ConvNets on CIFAR10 dataset.}
\label{tab:vae_gan}
\end{table}

\subsection{Comparison to MMD Baseline}
Another baseline is to learn synthetic images by distribution matching with vanilla MMD in the pixel space. This baseline can also been considered as the ablation study of the embedding function and differentiable  augmentation in our method.
We try this baseline with linear, polynomial, RBF and Laplacian kernels and with various kernel hyper-parameters. We find that only MMD with linear kernel can achieve better synthetic images, \ie better than randomly selected real images. The performance of MMD with linear kernel in the pixel space is presented in \Cref{tab:vae_gan}, which outperforms all generative models while is inferior to our method. This result also verifies that the distribution matching mechanism enables learning more informative synthetic samples.

\subsection{Comparison to GTN and KIP Methods}
We notice the recent works Generative Teaching Networks (GTN) \cite{such2020generative} and Kernel Inducing Point (KIP) \cite{nguyen2021dataset, nguyen2021dataset2} on dataset condensation. Such \emph{et al.} \cite{such2020generative} propose to learn a generative network that outputs condensed training samples by minimizing the meta-loss on real data. They report the performance of 4,096 synthetic images learned on MNIST which is comparable to our 50 images/class synthetic set (\ie 500 images in total) performance. 

Nguyen \emph{et al.} \cite{nguyen2021dataset, nguyen2021dataset2} propose to replace the neural network optimization in the bi-level optimization \cite{wang2018dataset} with kernel ridge regression which has a closed-form solution. Zero Component Analysis (ZCA) \cite{krizhevsky2009learning} is applied for pre-processing images. Although Nguyen \emph{et al.} \cite{nguyen2021dataset2} report the results on 1024-width neural networks while we train and test 128-width neural networks, our results still outperform theirs in some settings, for example $98.6\pm0.1\%$ v.s. $98.3\pm0.1\%$ when learning 50 images/class on MNIST and $29.7\pm0.3\%$ v.s. $28.3\pm0.1\%$ when learning 10 images/class on CIFAR100. 
Note that they achieve those results by leveraging distributed computation environment and training for thousands of GPU hours. In contrast, our method can learn synthetic sets with one GTX 1080 GPU in dozens of minutes, which is significantly more efficient. 

\section{Extended Visualization and Analysis}
\begin{figure*}[h]
    \centering
    \includegraphics[width=1\linewidth]{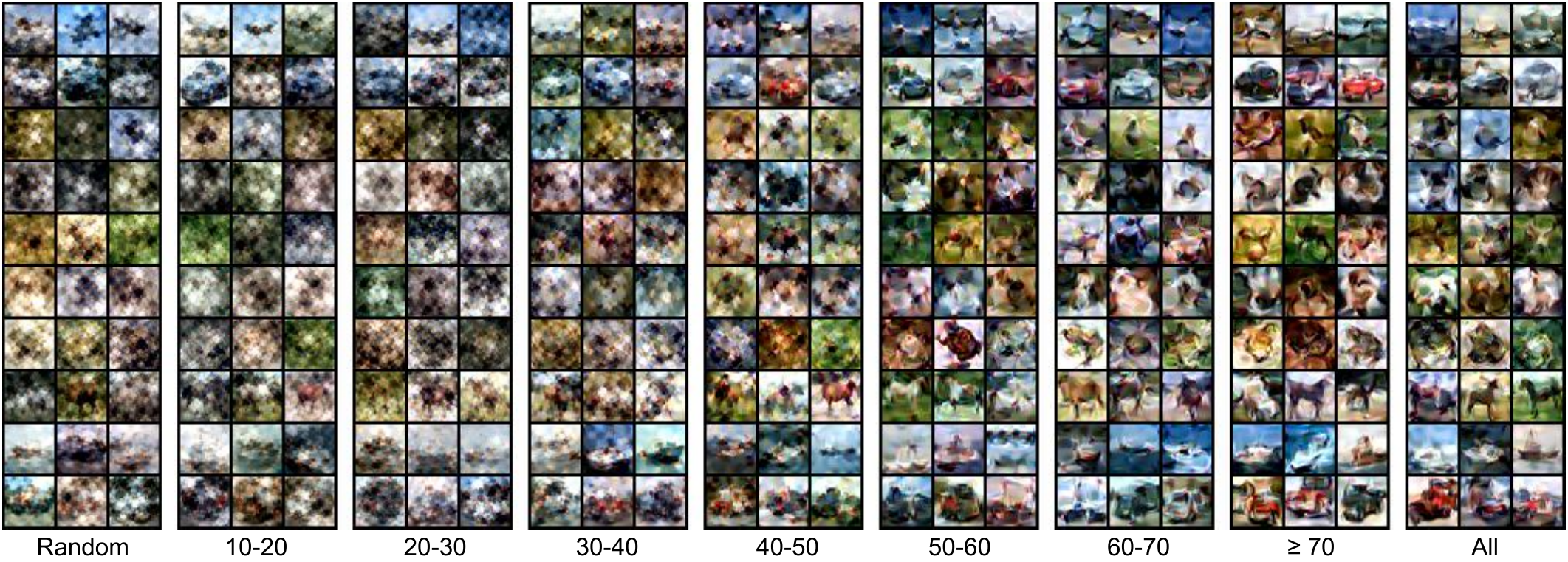}
    \caption{{Synthetic images of CIFAR10 dataset learned with different network parameter distributions, \ie networks with different validation accuracies (\%). Each row represents a class.}}
    \label{fig:vis_net_distribution}
\end{figure*}

We visualize the 10 images/class synthetic sets learned on CIFAR10 dataset with different network parameter distributions in \Cref{fig:vis_net_distribution}. It is interesting that images learned with ``poor'' networks that have lower validation accuracies look blur. We can find obvious checkerboard patterns in them. In contrast, images learned with ``good'' networks that have higher validation accuracies look colorful. Some twisty patterns can be found in these images. Although synthetic images learned with different network parameter distributions look quite different, they have similar generalization performance. We think that these images are mainly different in terms of their background patterns but similar in semantics. It means that our method can produce synthetic images with similar network optimization effects while significantly different visual effects. Our method may have promising applications in protecting data privacy and federated learning \cite{lyu2020threats}.

\section{Connection to Gradient Matching}
In this section, we show the connection between gradient matching \cite{zhao2021DC} and our method. 
Both \cite{zhao2021DC} and our training algorithm sample real and synthetic image batches from one class in each iteration, which is denoted as class $y$.
We embed each training sample $(\bx_i, y)$ and obtain the feature $\be_i$ using a neural network $\psi_{\bvartheta}$ followed a linear classifier $\mathbf{W} = [\bw_0, ..., \bw_{C-1}]$, where $\bw_j$ is the weight vector connected to the $j^\text{th}$ output neuron and $C$ is the number of all classes. 
Note that the weight and its gradient vector are organized in the same way in \cite{zhao2021DC}. We focus on the weight and gradient of the linear classification layer (\ie the last layer) of a network in this paper. 
The classification loss $J_i$ of each sample is denoted as
\begin{equation}
\begin{split}
    J_i = & - \log \frac{\exp{(\bw_y^T \cdot \be_i)}}{\Sigma_k \exp{(\bw_k^T \cdot \be_i)}}. \\
\end{split}
\end{equation}
Then, we compute the partial derivative w.r.t. each weight vector,
\begin{equation}
    \begin{split}
    \bg_{i,j} = \frac{\partial J_i}{\partial \bw_j} = & \left\{
    \begin{aligned}
        - \be_i + \frac{\exp{(\bw_y^T \cdot \be_i)}}{\Sigma_k \exp{(\bw_k^T \cdot \be_i)}} \cdot \be_i, \;\;\; j = y \\
        \frac{\exp{(\bw_j^T \cdot \be_i)}}{\Sigma_k \exp{(\bw_k^T \cdot \be_i)}} \cdot \be_i, \;\;\; j \neq y 
    \end{aligned}
    \right.
    \end{split}
\end{equation}
This equation can be simplified using the predicted probability $p_{i,j} = \frac{\exp{(\bw_j^T \cdot \be_i)}}{\Sigma_k \exp{(\bw_k^T \cdot \be_i)}}$ that classifies sample $\bx_i$ into category $j$:
\begin{equation}
    \begin{split}
    \bg_{i,j} = \left\{
    \begin{aligned}
        (p_{i,y}-1) \cdot \be_i, \;\;\; j = y \\
        (p_{i,j}-0) \cdot \be_i, \;\;\; j \neq y 
    \end{aligned}
    \right.
    \end{split}
\label{eq.connection_g_f}
\end{equation}

Eq. \ref{eq.connection_g_f} shows that \textbf{the last-layer gradient vector $\bg_{i,j}$ is equivalent to a weighted feature vector $\be_i$ and vice versa}. The weight is a function of classification probability.  Generally speaking, the weight is large when the difference between predicted probability $p_{i,j}$ and ground-truth one-hot label ($1$ or $0$) is large.

As the real and synthetic samples in each training iteration are from the same class $y$, we can obtain the mean gradient over a data batch by averaging the corresponding gradient components:
\begin{equation}
    \begin{split}
        \frac{1}{N} \Sigma_i^N \bg_{i,j} = \left\{
    \begin{aligned}
        \frac{1}{N} \Sigma_i^N (p_{i,y}-1) \cdot \be_i, \;\;\; j = y \\
        \frac{1}{N} \Sigma_i^N (p_{i,j}-0) \cdot \be_i, \;\;\; j \neq y 
    \end{aligned}
    \right.
    \end{split}
\end{equation}
$N$ is the batch size.
\textbf{Thus, last-layer mean gradient is equivalent to the weighted mean feature, and the mean gradient matching is equivalent to the matching of weighted mean feature.}

Our method can learn synthetic images with randomly initialized networks. Given networks with random parameters, we assume that the predicted probability is uniform over all categories, \ie $p_{i,j} = \frac{1}{C}$.
Then, the mean gradient is 
\begin{equation}
    \begin{split}
        \frac{1}{N} \Sigma_i^N \bg_{i,j} = & \left\{
    \begin{aligned}
        \frac{1-C}{C} \cdot \frac{1}{N} \Sigma_i^N \be_i, \;\;\; j = y \\
        \frac{1}{C} \cdot \frac{1}{N} \Sigma_i^N \be_i, \;\;\; j \neq y 
    \end{aligned}
    \right.
    \end{split}
\end{equation}
which is equivalent to the mean feature with a constant weight. 
\textbf{Thus, with randomly initialized networks, the last-layer mean gradient matching is equivalent to mean feature matching multiplied by a constant.}

\end{document}